# Classification of Smoking and Calling using Deep Learning

Miaowei Wang, Alexander William Mohacey, Hongyu Wang, James Apfel
{miaoweiw, amohacey, wahongyu, japfel}@umich.edu

## Executive Overview

*Since 2014, very deep convolutional neural networks have been proposed and become the must-have weapon for champions in all kinds of competition. In this report, a pipeline is introduced to perform the classification of smoking and calling by modifying the pretrained inception V3. Brightness enhancing based on deep learning is implemented to improve the classification of this classification task along with other useful training tricks. Based on the quality and quantity results, it can be concluded that this pipeline with small biased samples is practical and useful with high accuracy.*

## 1. Background and Impact

Code of conduct, that is some special restrictions of people's behavior in certain public scenes such as no smoking in gas stations, no calling for drivers, no taking photos in museums, etc. In fact, these behaviors have been causing huge disasters. 2.5 million nonsmoker adults have died because they breathed secondhand smoke since the 1965 Surgeon's Report [1]. Concretely, smoking during pregnancy results in more than 1,000 infants deaths annually; 34,000 premature deaths have been caused by secondhand smoke from heart disease each year in the United States among nonsmokers; Secondhand smoke causes more than 7,300 lung cancer deaths among U.S. nonsmokers each year [2]. Besides, calling in inappropriate environments also results in some irreparable damage. The National Safety Council reports that cell phone use while driving leads to 1.6 million crashes each year and nearly 390,000 injuries occur each year from accidents caused by texting and driving. In fact 94 percent of drivers support a ban on texting while driving. All in all, a detection system is urgently needed to determine whether people perform such dangerous behavior in certain environments to help reduce the accidents and death caused by smoking and calling.This report will introduce a practical pipeline to finish the behavior classification of smoking,calling ,normal and smoking-calling. In order to achieve high accuracy, we focus on deep learning methods with very deep convolutional neural networks such as inception V3. Since 2014,very deep convolutional neural networks have been proposed and become the must-have weapon for champions in all kinds of competition. This project tries to assess classification methods quantitatively and qualitatively. Since manual check for this dangerous behavior is laboursome and expensive, if such a system is deployed in a variety of public places that can be monitored, people who smoke will be given a warning and drivers will realize a risky behavior they do just now. In this way, second smoking will be reduced and help build a fresh environment to improve the quality of life and accidents will also be avoided when a system tries to hold back the calling behavior of some drivers.

## 2. Method

Classification based on deep learning achieves high accuracy and scores in many competitions. Very deep and complicated convolutional neural networks have been proposed during recent several years and received wide attention and recognition. Next, this section is going to present the main classification network framework, inception V3 which was trained using a dataset of 1,000 classes by google from the original ImageNet dataset which was trained with over 1 million training images. Inception v3 is a widely-used image recognition model that has been shown to attain greater than 78.1% accuracy on the ImageNet dataset. The model is the culmination of many ideas developed by multiple researchers over the years. It is based on the original paper: "Rethinking the Inception Architecture for Computer Vision" by Szegedy, et al[3].

The model itself is made up of symmetric and asymmetric building blocks including convolutions, average pooling max pooling, concats, dropout and fully connected layers. Batchnorm is used extensively throughout the model and applied to activation inputs.Loss is computed via Softmax. A high-level diagram of the model is shown below in FIgure 1. The Inception V3 model has approximately 25 million parameters, and it takes 5 billion multiplicative instructions to classify an image. On a modern PC without the GPU, classifying an image can be done in an instant.

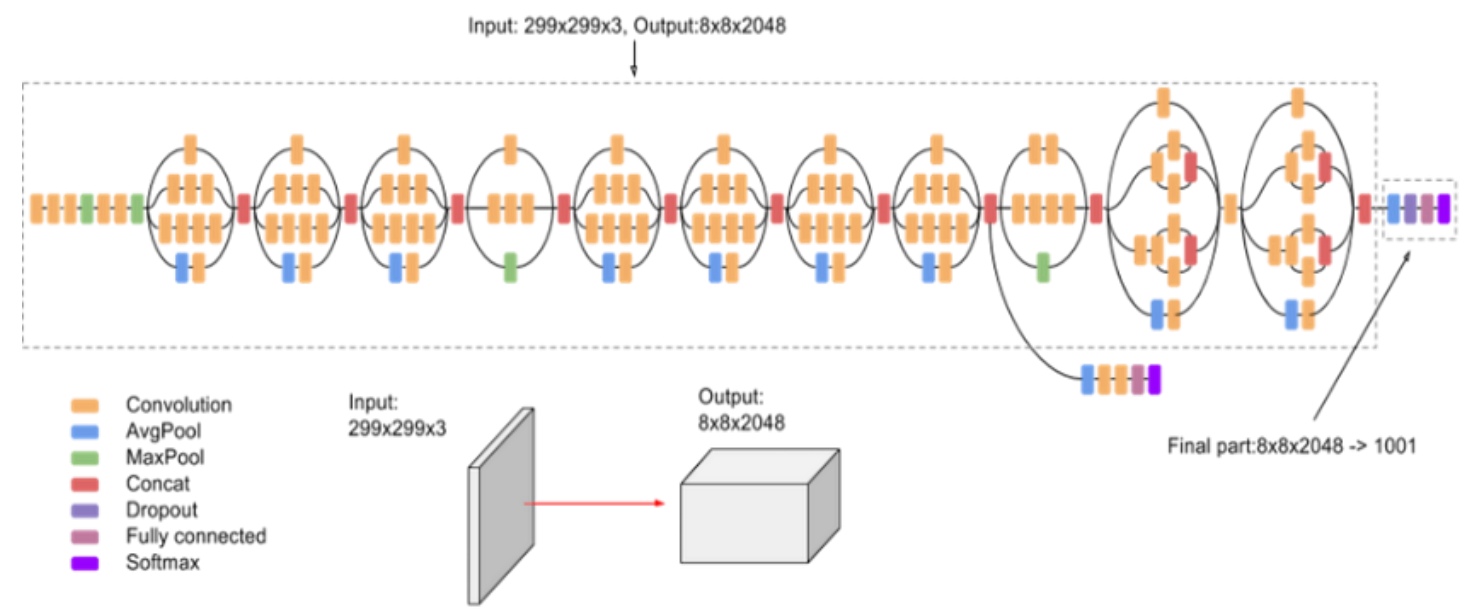

Figure1. InceptionV3 architecture diagram

Several lightings about inception V3 are worth further paraphrasing: for one layer of the neural network, more active output branches can generate mutually decoupled feature representation, thus generating high-order sparse features and accelerating convergence; reasonable use of dimension reduction does not destroy the network characteristic representation ability but can accelerate the convergence speed; an n*n convolution kernel can reduce dimensions by connecting two 1*n and n*1 in sequence which is similar to the matrix factorization; For multi-classification sample labeling, it is generally one-hot, such as [0,0,0,1]. The loss function similar to cross entropy will make the probability of the ground truth label assigned too much confidence in model learning. Moreover, due to the large gap between the logit value of ground truth labels and other labels, overfitting will occur, resulting in the reduction of generalization. One solution is to add regular terms, that is, to adjust the probability distribution of the sample label and make the sample label "soft", such as [0.1,0.2,0.1,0.6]. In this way, the error rate of top-1 and top-5 is reduced by 0.2% in the experiment.

As a useful image augmentation tool, this project also explores the up-to-date illuminance enhancement algorithm. Kindling the Darkness [4] proposed by Yonghua et is indeed useful for the image enhancing. The architecture of KinD(Kinding the Darkness) network is shown in Figure2.

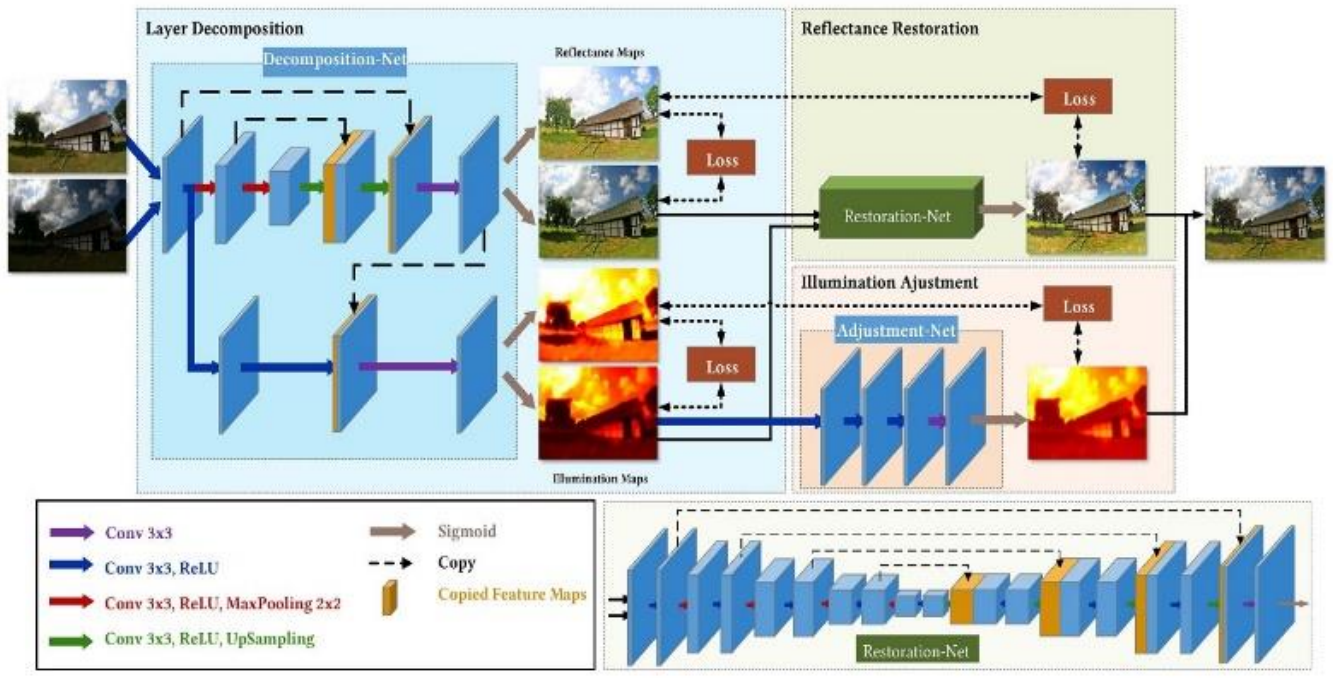

Figure2. The architecture of KinD network

In Figure 2 two branches correspond to the reflectance and illumination, respectively. From the perspective of functionality, this network can be divided into three subnets: 1. Layer Decomposition, which is same as retinex theory. In this net, reflectance graph of high light image can be used as a ground truth to lead enhancement of low light image's reflectance graph. Of course, there is no real ground truth, a bright image is just a high-light input Figure;2. Illumination Guided Reflectance Restoration: the goal of this part is to restore the contaminated part by means of Illumination map guidance, since the retinex formula is related to Illumination map;3. Arbitrary Illumination Manipulation: a flexible mapping mechanism is proposed, which is derived from the image data and, of course, allows users to manually adjust the mapping.

# 3. Prototype

## 3.1 Datasets

This project's dataset comprises of four classes, smoking behavior(1150 images), normal behavior(2625 images), calling behavior(2025 images), and smoking-calling behavior(1400 images).The dataset is provided by China eHualu group Co. ,Ltd. who crawls the data from the whole network for non-profit use[5]. Some images of the dataset have been shown as Figure3 below. From the Figure3, the dataset is really difficult and complicated with all kinds of resolutions, different illumination, rotated and distance varying images, and a variety of sceneries. Once this project gets high accuracy in such tanglesome images, it is proved that the pipeline has powerful robustness and portability.

## 3.2 Training Tricks and Implementation

This program is built based on an open-source project doing classification of marine fish species [6] for fisheries regulation using keras deep learning API. This project improves the overall organization and flow. As the original code used deprecated code, this project adjusted some features (notably updating Model.fit_generators to Model.fit) in order to improve performance. This also allowed us to use some features that would greatly simplify other parts of the code: for example, the new model fit function supported native learning rate schedulers, such as keras.experimental.CosineDecay.

The other major flow improvement and code simplification was using the built-in validation_split feature of the ImageDataGenerator preprocessor, which allows Keras to automatically split the data while building the data generators, without the need for a separate script. The only preprocessing implemented within the Keras image generators was flipping images horizontally at random (to reduce right/left-handed bias in our set). The data still supports external preprocessing. The Model.fit function supports automatically calculating epoch steps based on batch size and total image count, which may result in misreported set sizes with irregular batch sizes. We used a relatively large initial learning rate, as the introduction of cosine decay means that we can use a much larger initial learning rate and let it drop over time, resulting in a reduction of training time by over a third. Finally, we implemented a callback function to automatically track the improvement of the model, and to stop training early if the validation accuracy starts to degrade, indicating that the model is starting to overtrain. Overall, these improvements and adjustments reduce the training process to a single script, down from an initial three, and dramatically reduce training time with no effective loss in final accuracy.

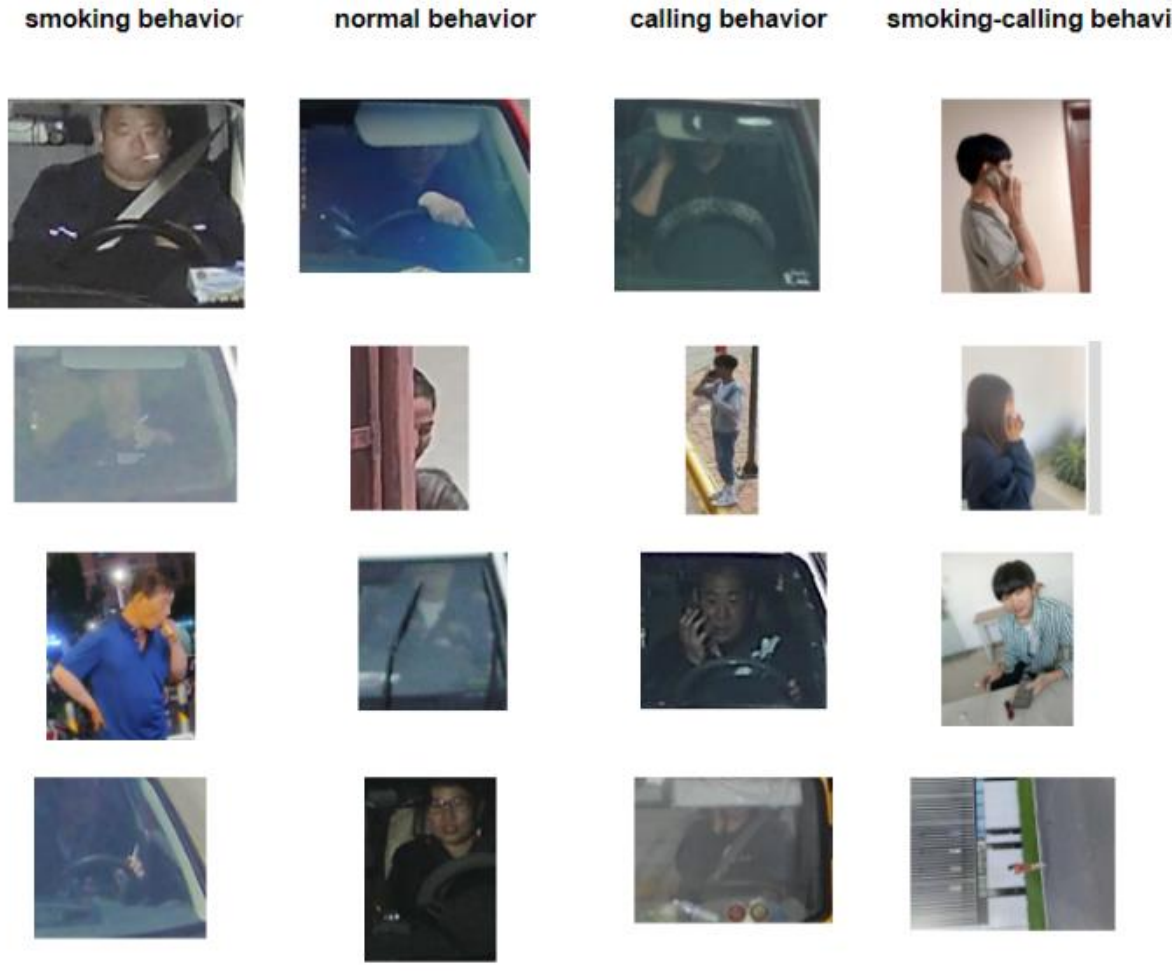

Figure3. Some samples from this project's training dataset

## 3.3 Image Enhancement

In order to increase the accuracy and efficiency of classification, this project tries to enhance the brightness of darker images to increase the accuracy of the classification results. To realize this function, this project has two steps to consider: (1) Find those relatively dark images. (2) Enhance the brightness of these images. For the first step, this project can calculate the proportion of pixels with lower brightness to all pixels. After several attempts, the project's researchers believe that pixels with grayscale values below 50 can be regarded as the "darker pixels". And if the proportion of darker pixels in an image is greater than 0.3, then it can be considered that this image needs a certain degree of brightness enhancement. The second step can be realized based on an open-source project on Github[7] which is an implement of the KinD network the method section has introduced. By the pre-trained checkpoints, the brightness of images can be enhanced. Two images with a dark pixel ratio of 0.3 and 0.6 is shown in Figure4(a) and the enhanced results of them are shown in Figure4(b).

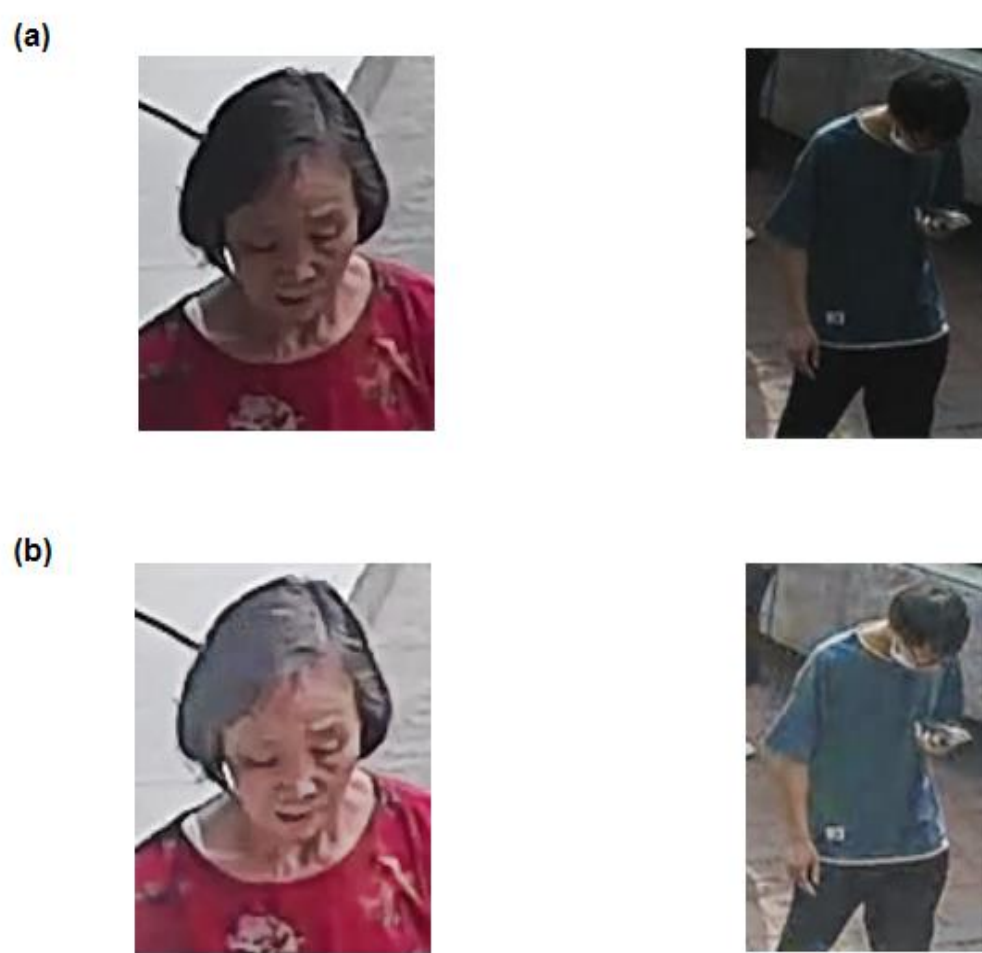

Figure4. Image Enhancement by KinD

Some statistics about the enhancing was noted down in Table1

Tabel1. Statics about enhancing

| | **Smoking** | **Normal** | **Calling-smoking** | **Calling** |
|---|---|---|---|---|
| Number of the original images | 2625 | 1150 | 1400 | 2025 |
| Number of dark images | 603 | 312 | 383 | 656 |
| Percentage of dark images | 23.0% | 27.1% | 27.4%1 | 32.4% |

## 4. Results

In training, this pipeline's accuracy and validation logs demonstrated nothing out of the ordinary, with this project's validation reaching on average a final accuracy of approximately 94% before ceasing improvement (see the charts in Figure 5 on the next page for reference). Initially, spurious preprocessing settings resulted in odd trends where the training accuracy and loss lagged far behind the validation accuracy; however, this was resolved by removing unnecessary preprocessing at training time.

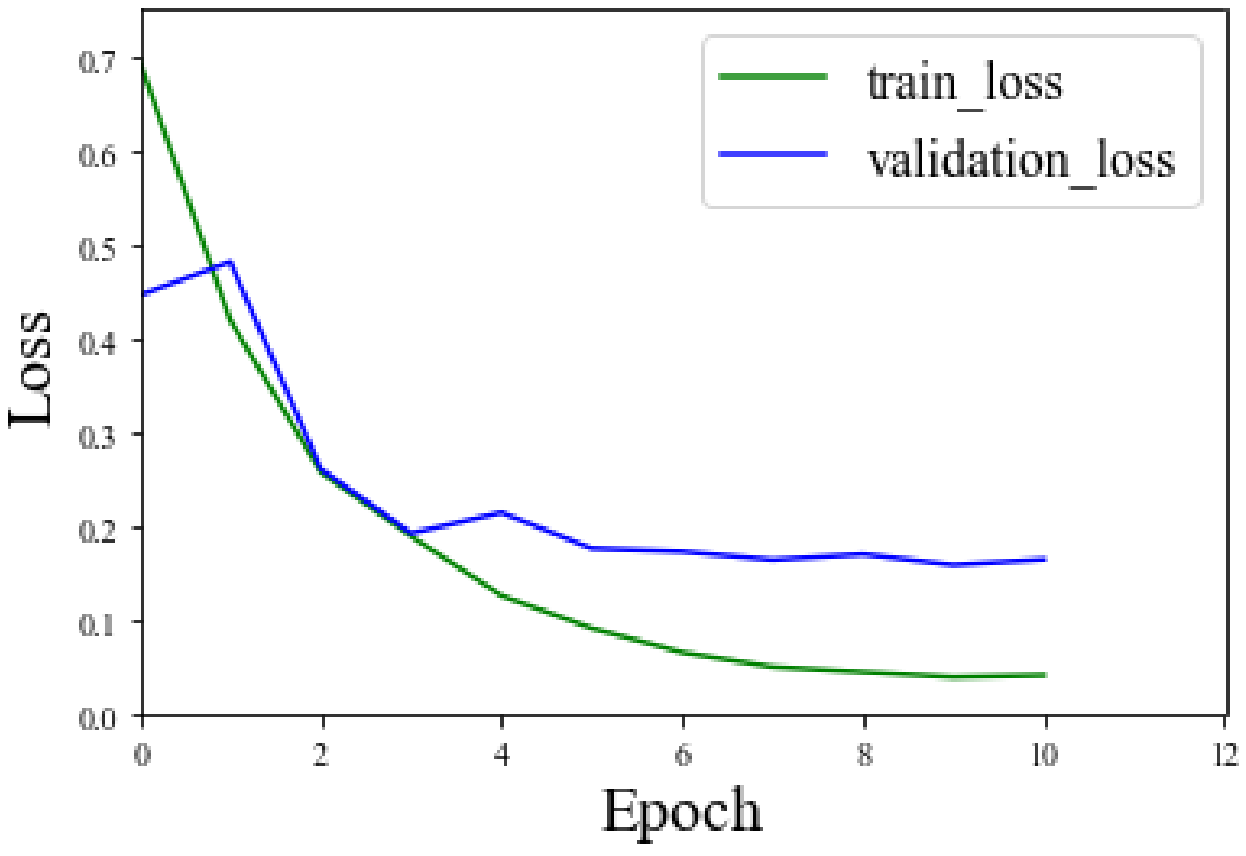

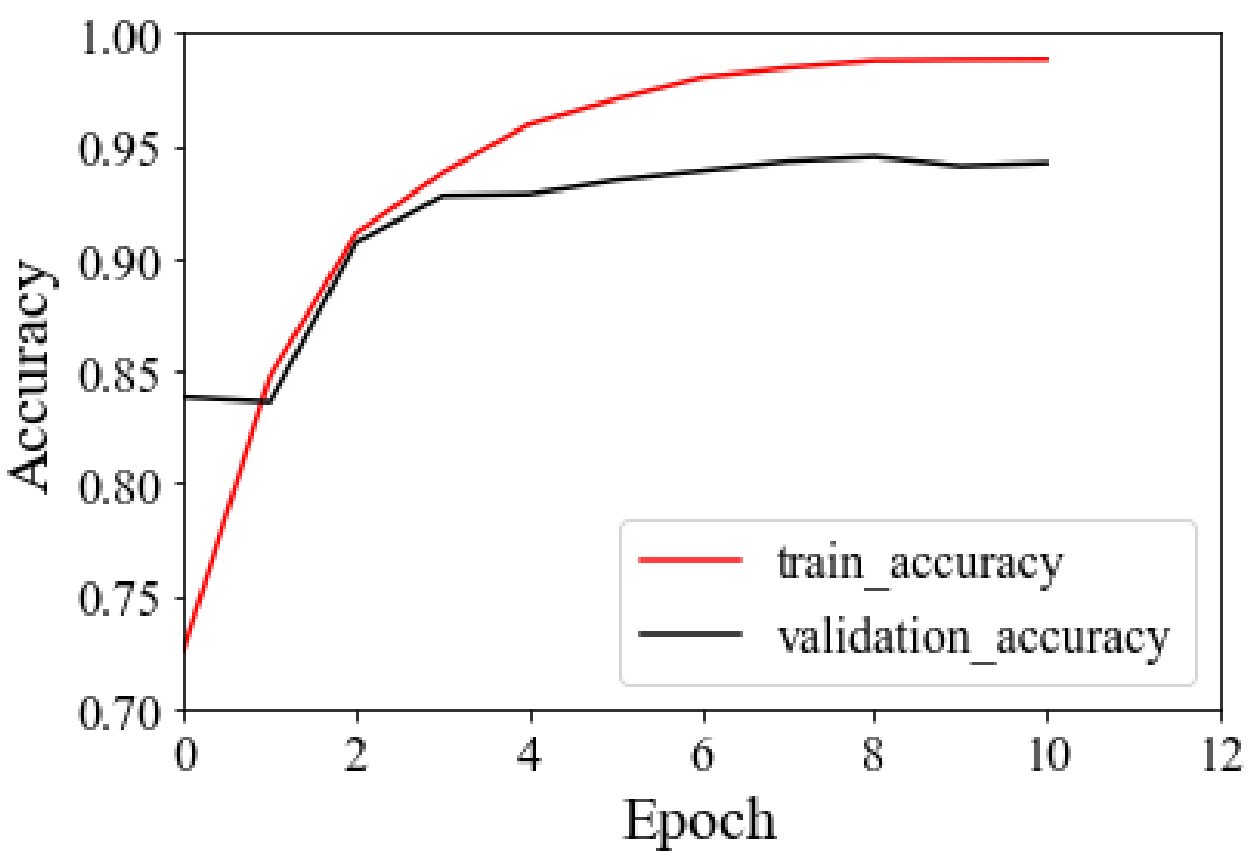

Figure 5. Accuracy and Loss vs Epoch

Image classification of individual images was executed in the image_loc.py script, using this project's trained model and the OpenCV 2 library. Test images may be of any format compatible with OpenCV and any resolution above approximately 150x150. OpenCV was used to downsize images to 299x299 for classification. Image classification was performed using the trained model and the resulting highest probability label was placed on the full sized image and exported as a PNG. A rudimentary localization of the label was then performed in the image by dividing the image into a 4x4 grid. Each of the 16 tiles was then classified individually, if the frame had the same classification as the overall image, the frame was highlighted and the probability printed to it. These tiles were then reconstructed into the original image and output again as shown in Figure 6. This method is 20x slower than processing a single image (0.2s vs 4s), as it involves expensive image cropping using OpenCV and running the model 16 additional times. Although the InceptionV3 model doesn't support object detection, the 8x8x2048 layer before pooling could be modified to do so. Localization for the 8x8 grid could be trained and the grid could be cut with an inter-cell L1 loss function, combining the results with non-maximum suppression to form bounding boxes. An alternative Keras based model to better implement localization is yoloV5, which supports object detection.

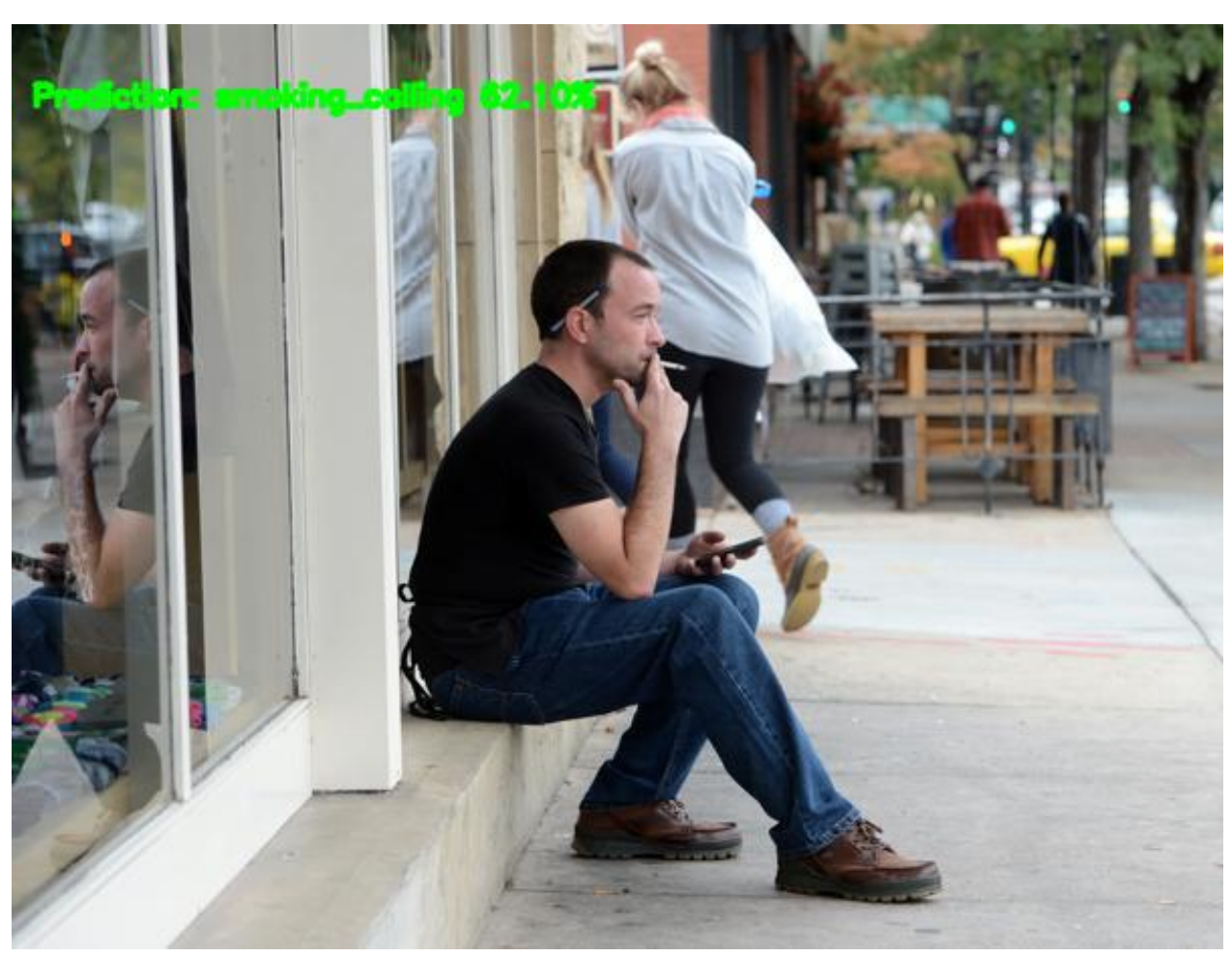

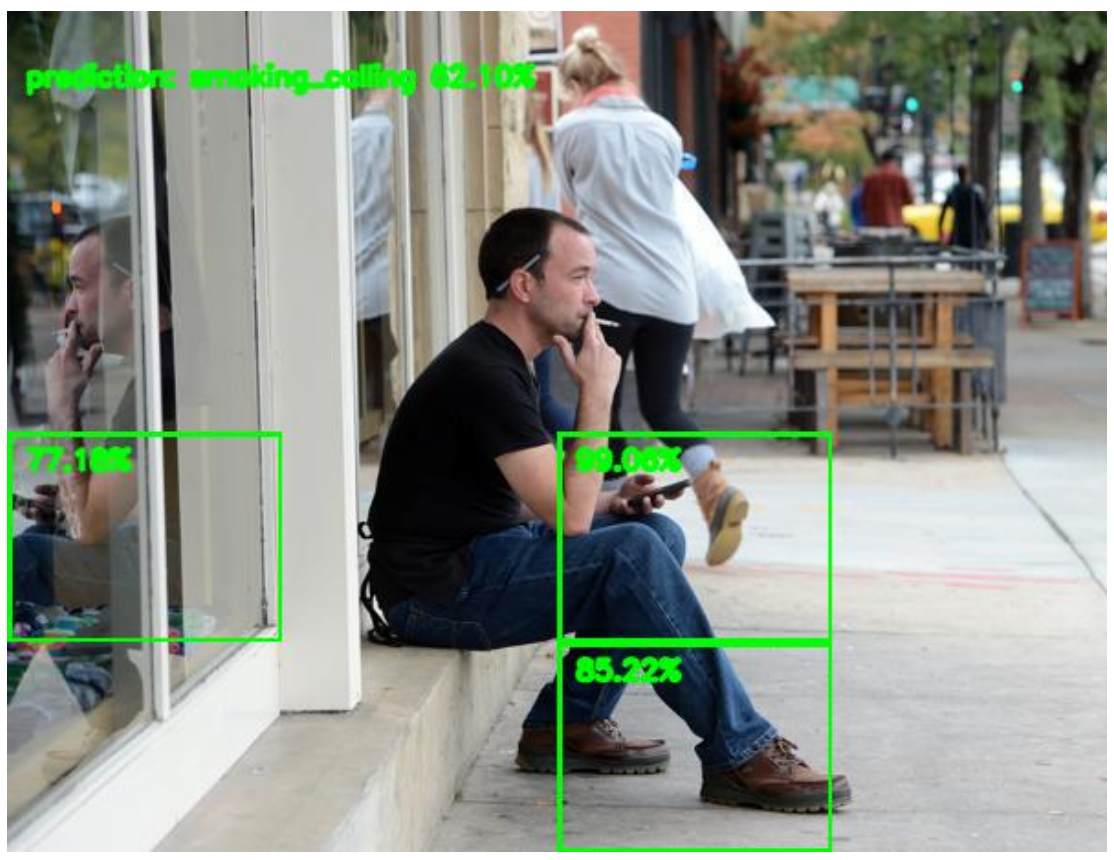

Figure 6. Whole image classification of an image and localization of matching image segments for smoking-calling behavior.

Video classification was executed in our videopredict.py script, again using this project's trained model and the OpenCV 2 library. Test videos used were encoded in the MP4 format, at a resolution of 720p and framerate of 30 fps. OpenCV was used to split the video into individual frames and downsized to 299x299 for classification. A floating mode of the prior 15 frames was used to select the most common image classification label. A floating mean of the floating mode's probabilities was then calculated. The classification information from the floating mode and mean were then placed on the frame and the frame was then output into the .avi file format via OpenCV. Screenshots of some resulting video frames are shown in Figure 7. Video accuracy was 81 percent for detecting smoking and 85 percent calling. Additional information printed for video runs included the number of frames processed, time to complete the processing, and percentage of the processed frames classified as each label. The average processing speed for an Intel i7-3770 CPU was 4.95 frames per second, which would be adequate for low FPS CCTV cameras for monitoring customers in a business or drivers. The percentage of frames processed that contain smoking could be used in an application such as automated film content warning ratings for smoking and tobacco use.

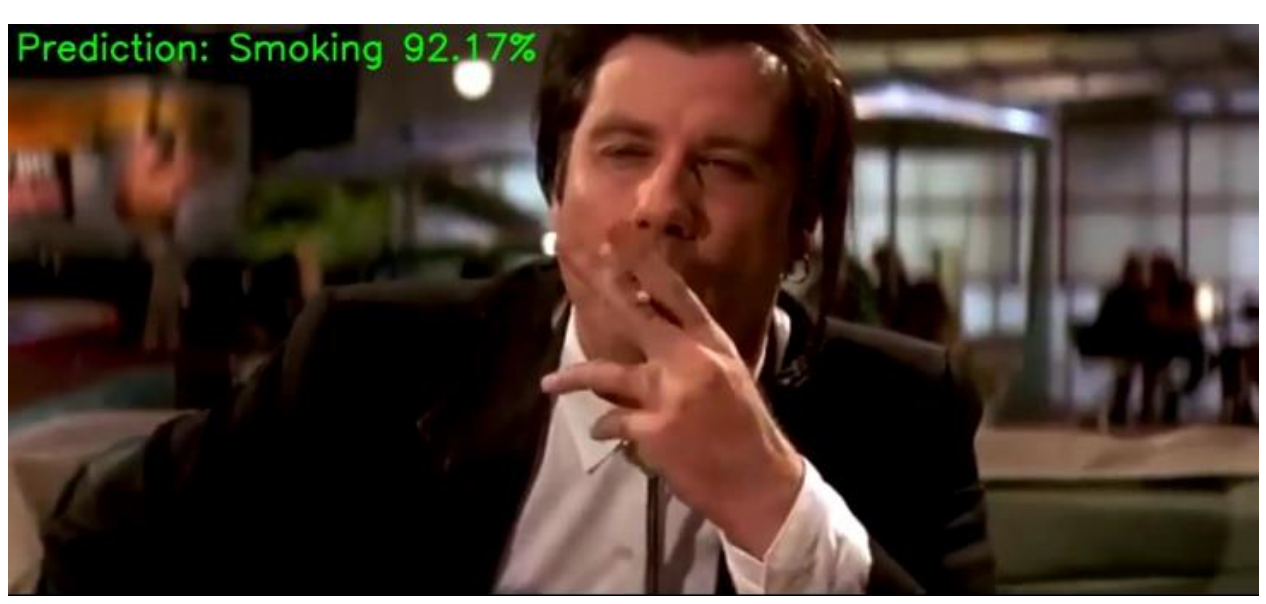

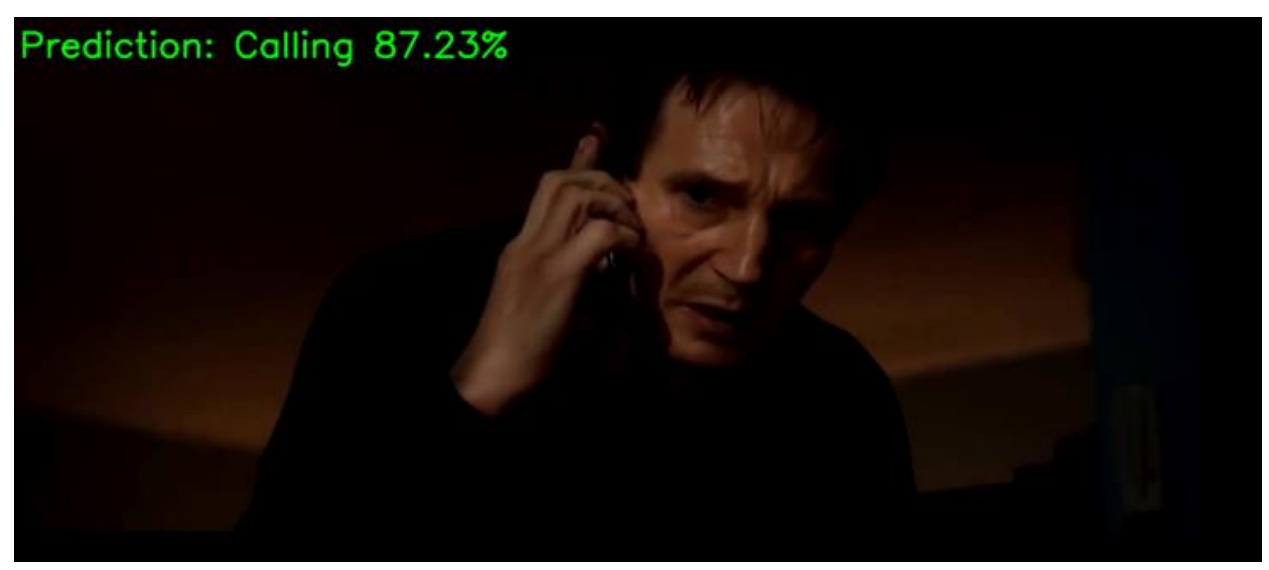

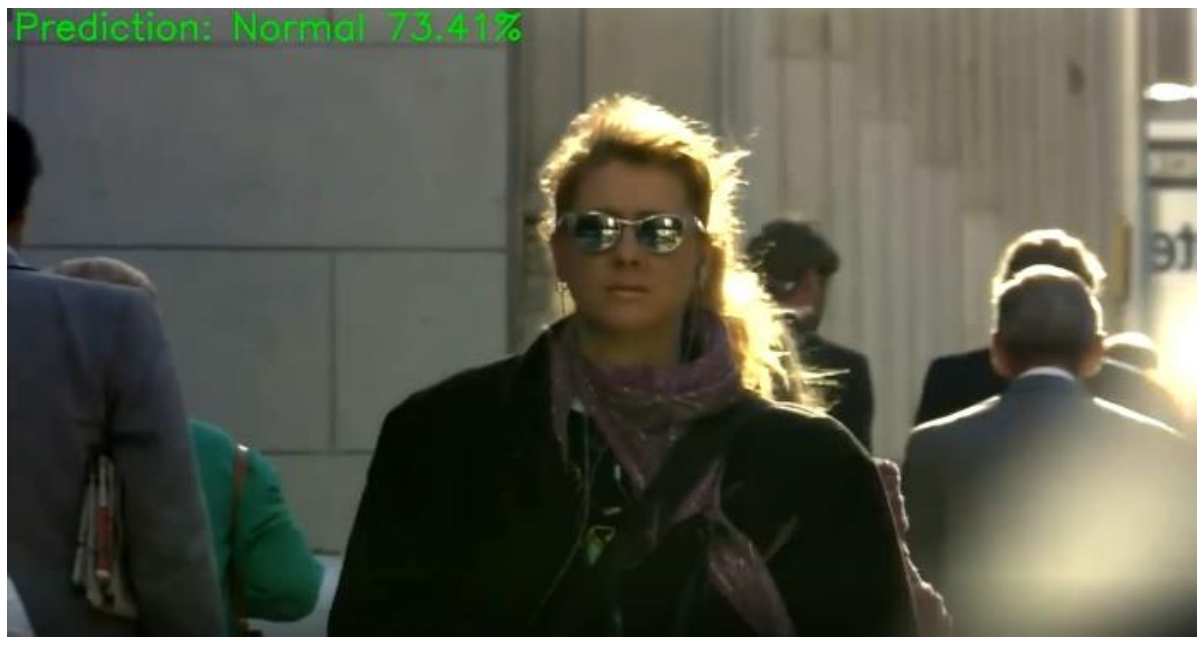

Figure 7.: Classification of video examples from the movies Pulp Fiction (1994) , Taken (2008), and a Sony Demo Tape (1993) for smoking, calling and normal behavior.

This project's source code is released in Github

https://github.com/wangmiaowei/CVDancer_Classification